\documentclass[runningheads]{llncs}
\usepackage[T1]{fontenc}
\usepackage{graphicx}
\usepackage{booktabs}
\usepackage[misc]{ifsym}
\newcommand{\corr}{(\Letter)}

\usepackage{multirow}
\usepackage{hyperref}

\begin{document}

\title{Machine Learning for Climate Policy: Understanding 
Policy Progression in the European Green Deal}

\titlerunning{Machine Learning for Climate Policy}

\author{Patricia West\inst{1}
\and
Michelle Wing Lam Wan\inst{1, 2}\orcidID{0000-0002-9090-2387}
\and
Alexander Hepburn\inst{1}\orcidID{0000-0002-2674-1478}
\and
Edwin Simpson\inst{1}\orcidID{0000-0002-6447-1552} 
\and
Raul Santos-Rodriguez\inst{1}\orcidID{0000-0001-9576-3905} 
\and
Jeffrey Nicholas Clark\inst{1}\orcidID{0000-0003-0118-3999} \corr}

\authorrunning{P. West et al.}

\institute{University of Bristol, UK \email{jeff.clark@bristol.ac.uk}
\and
University of Cambridge, UK 
}

\maketitle              

\begin{abstract}
Climate change demands effective legislative action to mitigate its impacts. This study explores the application of machine learning (ML) to understand the progression of climate policy from announcement to adoption, focusing on policies within the European Green Deal. We present a dataset of 165 policies, incorporating text and metadata. We aim to predict a policy's progression status, and compare text representation methods, including TF-IDF, BERT, and ClimateBERT. Metadata features are included to evaluate the impact on predictive performance. On text features alone, ClimateBERT outperforms other approaches (RMSE = 0.17, R$^2$ = 0.29), while BERT achieves superior performance with the addition of metadata features (RMSE = 0.16, R$^2$ = 0.38). Using methods from explainable AI highlights the influence of factors such as policy wording and metadata including political party and country representation. These findings underscore the potential of ML tools in supporting climate policy analysis and decision-making.

\end{abstract}

\section{Introduction}
The urgency of climate change necessitates robust policies and legislation to support adaptation and mitigation strategies, as well as an understanding of how policies progress through legislative processes. Time-sensitive identification of delayed or progressing policies can empower policy advocates to act strategically. However, policy texts are often long and complex, making manual analysis labour-intensive. Machine learning offers a scalable solution to analyse and understand policy progression, enhancing transparency and supporting policy advocacy.

With the emergence of large-language models, machine learning is increasingly being utilised to summarise and extract features from text documents~\cite{liu2019text}, such as legal documents~\cite{kanapala2019text}. Explainability techniques are often employed to increase trust and confidence in these systems~\cite{lyu2024towards}, which is key when dealing with sensitive documents. Previously, tools have been used to gather and analyse large datasets of EU legal documents, such as Judict~\footnote{\url{https://judict.eu/en}}, a database and analysis tool for documents relating to EU Financial law. Judict is a contribution to the 
\href{https://data.europa.eu/en}{data.europa.eu} datasets, which encourages reuse of European open data to address issues central to the EU. Unlike prior work, we focus on understanding the progression of environmental legislation.

This paper builds on the methodology proposed by Clark et al.~\cite{clark2023}, to gather data relating to EU climate policy and predict policy progression, enabling improved understanding of the policy adoption process. We aim to:

\begin{enumerate}
    \item Construct a novel dataset of climate policies from the European Green Deal, a cornerstone of the EU’s climate strategy.
    \item Apply machine learning techniques to predict policy progression categories.
    \item Use explainability techniques to investigate the role of text and metadata features in influencing predictions.
\end{enumerate}

\section{Methods}
Human-driven policy tracking can be effective at small scales, but becomes inefficient with higher volumes of policy documents. We therefore leverage language models to develop a scalable framework for the analysis of these documents, to complement existing expertise. 

\begin{figure}[t]
    \centering
    \includegraphics[width=1\linewidth]{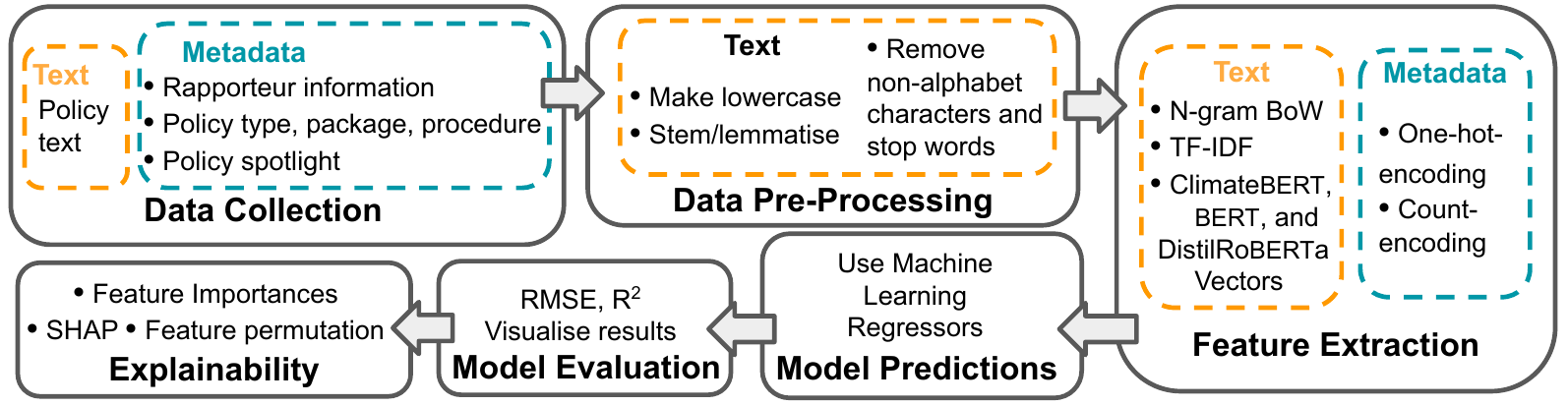}
    \caption{Processing pipeline, from policy and metadata data collection to category prediction and associated explanations.}
    \label{fig:paper_pipeline}
\end{figure}

\begin{figure}[t]
    \centering
    \includegraphics[width=1\linewidth]{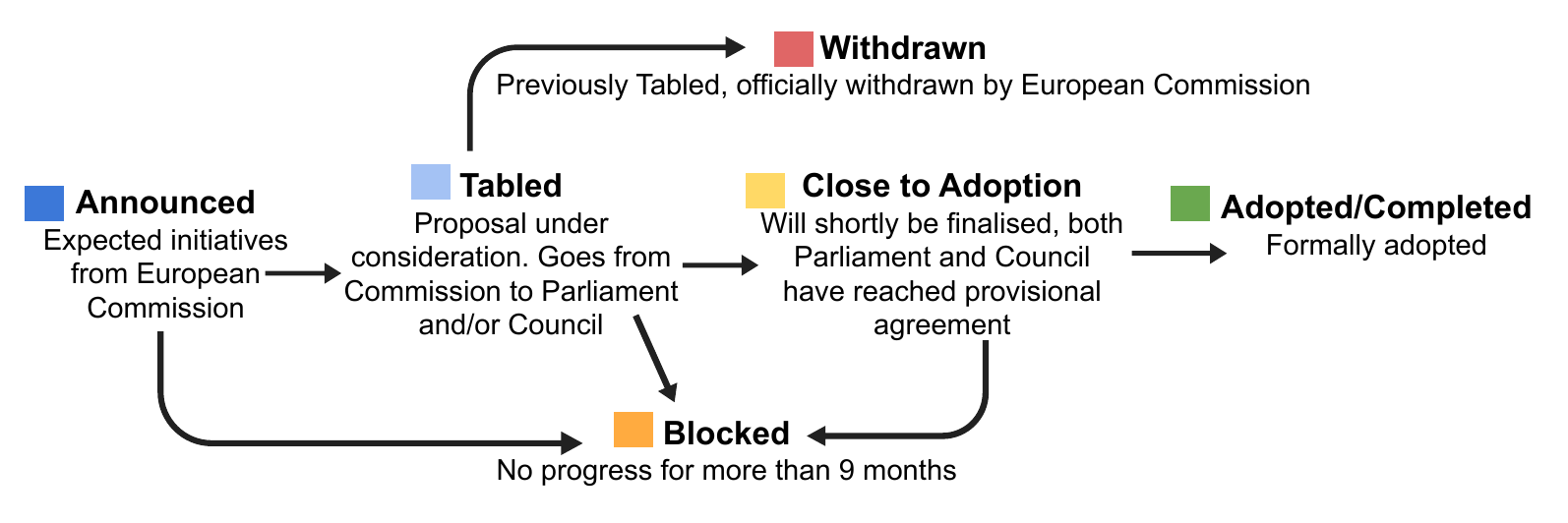}
    \caption{Categories of policy stages considered, in the path towards adoption.}
    \label{fig:paper_category_order}
\end{figure}

\begin{figure}[t]
    \centering
    \includegraphics[width=1\linewidth]{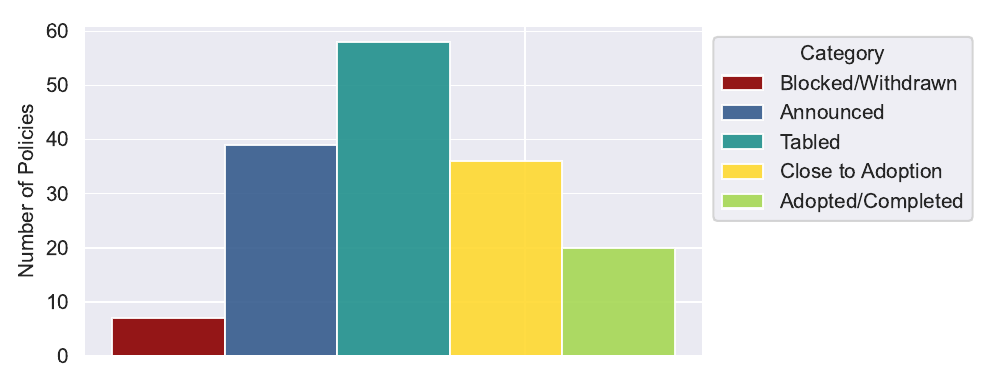}
    \caption{Distribution of policy stage labels in the dataset.}
    \label{fig:ap_labels}
\end{figure}

\paragraph{Data}
Our dataset consists of 165 policies, collected from the European Green Deal legislative tracker, European Train \cite{European_train}. All policies available before 29th January 2024 were included in this analysis. Our data processing pipeline can be found in Fig.~\ref{fig:paper_pipeline}. The six categories for the current status of each policy ranged from ``Announced'' to ``Adopted/Completed'' (Fig.~\ref{fig:paper_category_order}). The distribution of categories can be found in Fig.~\ref{fig:ap_labels}. Alongside policy text, 62 metadata features were gathered, including month and year of the policy, rapporteur information, policy type, and legislative procedure (Table ~\ref{tab:feat_summary}).
Text data was cleaned by removing non-alphabet characters and stop words. Tokenisation and lemmatisation ensured consistency. Metadata features, such as rapporteur country and party, were encoded.

\begin{table}[h!] 
\caption{Summary of metadata features included, in addition to the policy text}
\centering
\begin{tabular}{@{}lll@{}}
\toprule
\textbf{Feature Category} & \textbf{Feature Values} & \textbf{No. of Features} \\ \midrule
\multirow{2}{*}{Policy Update Date} & Month & 1 \\
 & Year & 1 \\ \midrule
\multirow{3}{*}{Rapporteur(s)' Country} & Country name (count encoded) & 20 \\
 & No rapporteur (binary) & 1 \\
 & Rapporteur country voting weight & 1 \\ \midrule
\multirow{3}{*}{Rapporteur(s)' Party} & Party name (one-hot encoded) & 7 \\
 & No party (binary) & 1 \\
 & MEP seat proportion & 1 \\ \midrule
\multirow{2}{*}{Spotlight} & Spotlight (binary) & 1 \\
 & Spotlight name (one-hot encoded) & 5 \\ \midrule
\multirow{2}{*}{Procedure} & Year & 1 \\
 & \begin{tabular}[c]{@{}l@{}}Procedure type (one-hot encoded) \end{tabular} & 4 \\ \midrule
Type & \begin{tabular}[c]{@{}l@{}}Legislative (binary) \end{tabular} & 1 \\ \midrule
\multirow{2}{*}{ClimateBERT} & Predicted score & 5 \\
 & Predicted label & 11 \\ \bottomrule
\end{tabular}
\label{tab:feat_summary}
\end{table}

\paragraph{Models}
We split policies across an 80:20 train:test split. Text features were represented using: (i) TF-IDF - emphasising distinctive terms; (ii) BERT - generating semantic embeddings \cite{bert}; (iii) ClimateBERT - additional finetuning of BERT on climate-related texts \cite{climatebert}. Regression models were applied to predict the current category of each policy, including CatBoost \cite{NEURIPS2018_14491b75}, Random Forest, Bayesian Ridge Regression, and Support Vector Regression (SVR)~\cite{hastie01statisticallearning}, to encompass models with diverse properties around linearity, interpretability or handling of categorical features. The target variable (policy category) was mapped onto a 0-1 scale to maintain the sequential nature: ``Withdrawn'' and ``Blocked'' = 0, ``Announced'' = 0.25, ``Tabled'' = 0.5, ``Close to Adoption'' = 0.75, ``Adopted/Completed'' = 1. Model performance was evaluated using RMSE and R$^2$ metrics, comparing the predicted and actual category for each policy. Although these are not standard metrics for classification tasks, they were chosen in order to emphasise the ordinal policy stages.

\paragraph{Explainability}
Permutation feature importance \cite{pedregosa2011scikit} and SHapley Additive exPlanation (SHAP) values \cite{shap} were utilised, to highlight influential features and validate feature contributions, with the aim of providing insights and understanding into the process of policy progression. All code for this work is available upon request.

\section{Results and Discussion}

\subsection{Model Performance}

For the task of predicting policy progression, highest performance was achieved using all available features, with representation by BERT combined with Bayesian Ridge Regression (RMSE = 0.16, R$^2$ = 0.38, Table \ref{tab:metadata_results}). When using only policy text features, representation by ClimateBERT combined with SVR provided best prediction performance  (RMSE = 0.17, R$^2$ = 0.29, Table \ref{tab:text_results}). In contrast to models using BERT for text feature representation, those using ClimateBERT demonstrated smaller performance improvements with the addition of metadata features. This may reflect the pretraining of ClimateBERT on climate texts, such that it was already capable of capturing relevant information from the policy texts without the inclusion of metadata. Unlike both BERT and ClimateBERT, the simpler TF-IDF text representation methods demonstrated inconsistent performance with additional metadata features, although their performance supports the importance of feature selection.


\begin{table}[h]
\caption{Results for policy category prediction across best-performing feature representation methods, regression models, when using both policy text and metadata features.}
\centering
\begin{tabular}{lllll}
    \toprule
    \textbf{Feature Representation} &  \textbf{Regression Model} & \textbf{RMSE} \(\downarrow \) & \textbf{R$^2$} \( \uparrow \)\\ \hline

    TF-IDF & Bayesian Ridge & 0.17 & 0.29 \\
    BERT & Bayesian Ridge & \textbf{0.16} & \textbf{0.38} \\
    BERT & SVR & \textbf{0.16} & 0.33 \\
    ClimateBERT & SVR & 0.17 & 0.32 \\
    \bottomrule
    
\end{tabular}

\label{tab:metadata_results}
\end{table}

\begin{table}[h]
\caption{Policy category prediction results using only policy text-based features and no metadata features.}
\centering
\begin{tabular}{lllll}
    \toprule

    \textbf{Feature Representation} & 
    \textbf{Regression Model} & 
    \textbf{RMSE} \(\downarrow \) & \textbf{R$^2$} \( \uparrow \)\\ \hline
    TF-IDF & Random Forest & 0.17 & 0.28 \\
    TF-IDF & CatBoost & 0.17 & 0.27 \\
    BERT & SVR & 0.17 & 0.26 \\ 
    ClimateBERT & SVR & 0.17 & \textbf{0.29} \\ 
    \bottomrule
\end{tabular}
\label{tab:text_results}
\end{table}

\subsection{Feature Contributions}
\begin{figure}
\centering
\includegraphics[width=\textwidth]{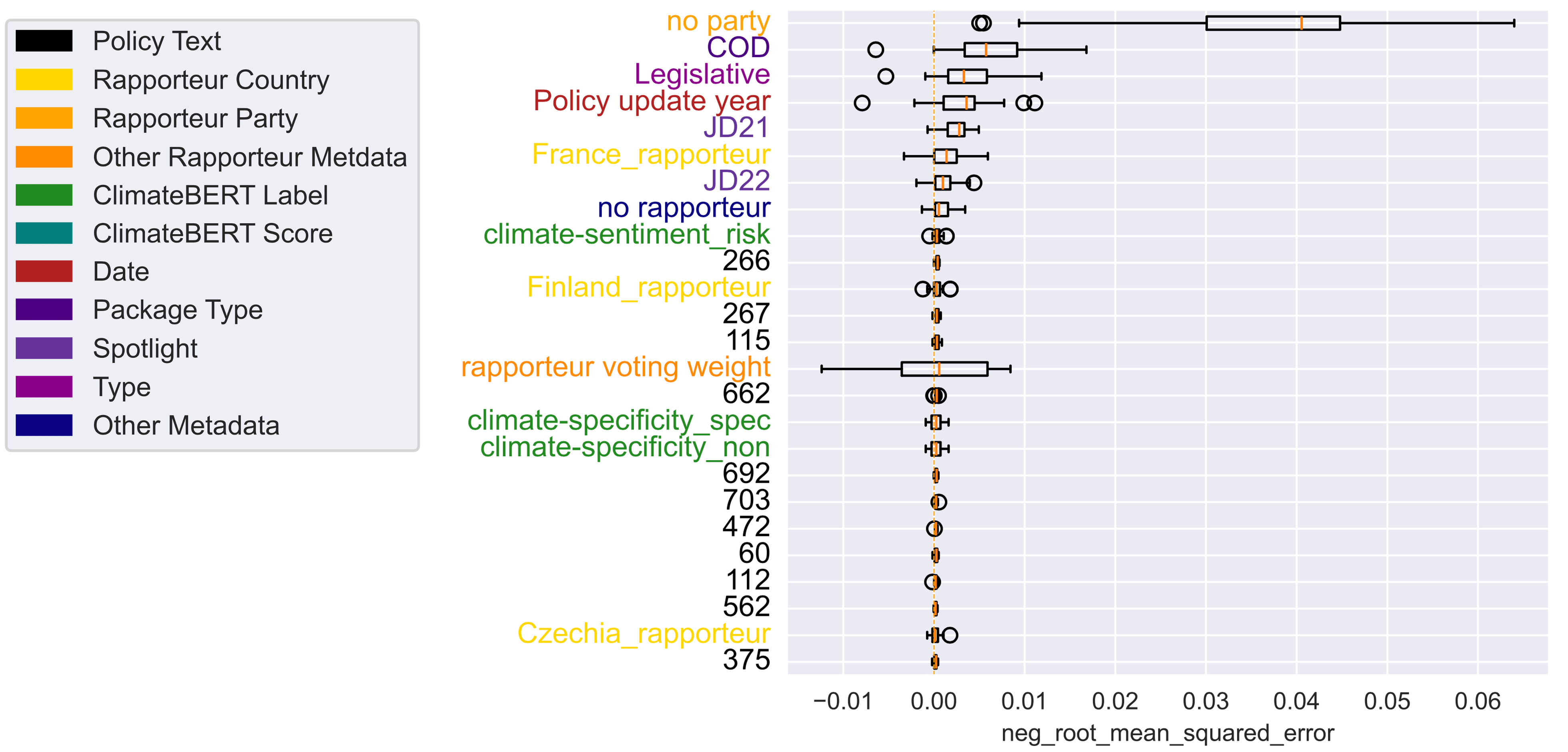}
\caption{\centering Feature permutation of BERT with Bayesian Ridge regression. Metadata features are colour-coded and text features are presented in black. Values are shown as the negative change in RMSE after shuffling each feature, so larger values indicate greater importance.}
\label{fig:BERT_feature_permutation}
\end{figure}

Explainability analysis for two approaches with text and metadata features are presented here. Firstly, Fig.~\ref{fig:BERT_feature_permutation} shows the results of feature permutation on the highest performing model: Bayesian Ridge regression model using BERT text features. The ``no party'' metadata feature value was observed as the most important feature to the prediction model, by a large margin. This feature represents policies which are not associated with a rapporteur, or have a rapporteur who is not associated with a major political party. The importance of this feature value may suggest confounding factors. Further work, with domain experts, should be carried out to assess these findings. Other metadata features highlighted by the feature permutation analysis included rapporteur country (France, Finland, Czechia), policy type (COD - ordinary legislative procedure; Legislative), and spotlight status (JD21, JD22 - EU legislative priorities for the year 2021 and 2022 respectively). 

\begin{figure}[h]
\centering
\includegraphics[width=\textwidth]{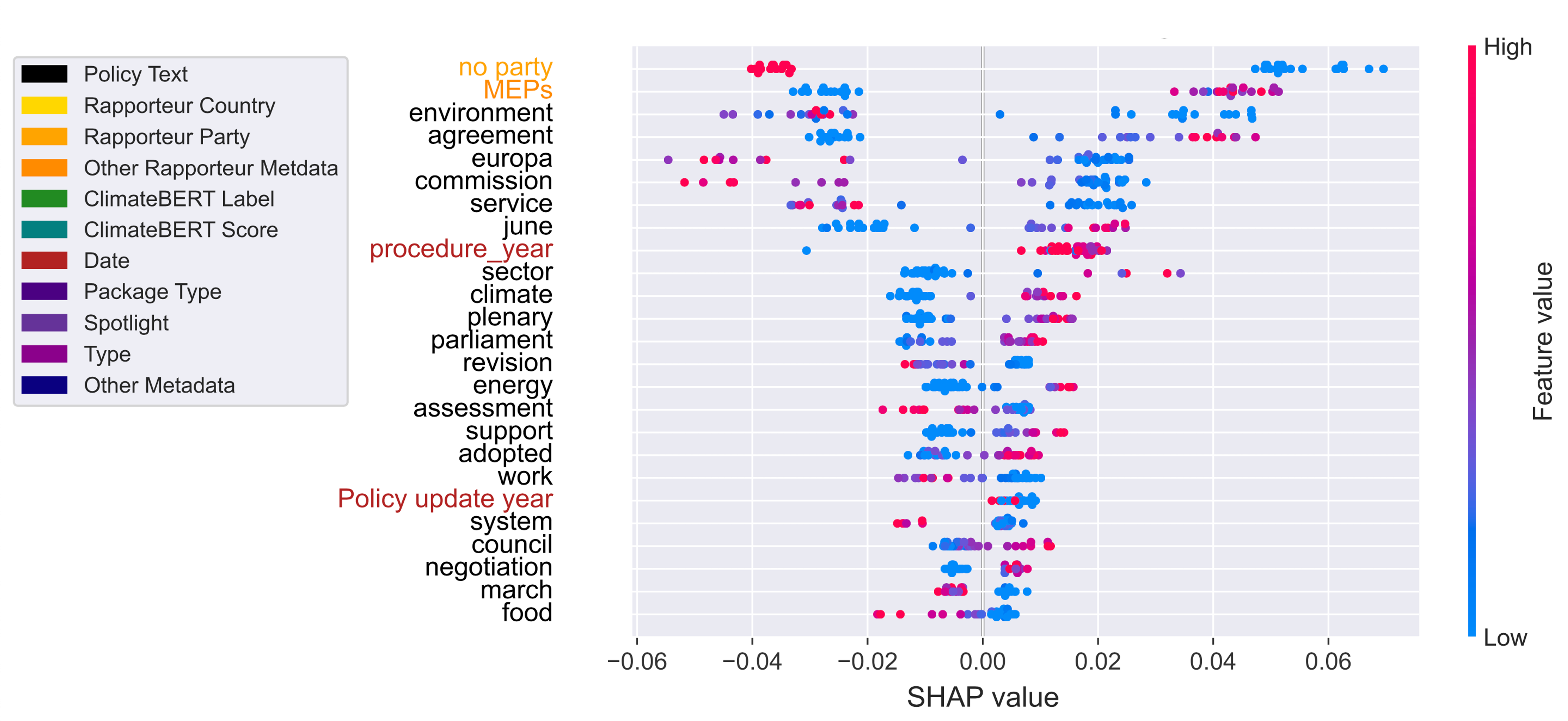}
\caption{\centering SHapley Additive exPlanation (SHAP) values of TF-IDF with CatBoost regression (RMSE = 0.18, R$^2$ = 0.2). Metadata features are colour-coded and text features are presented in black. Feature values are colour-coded from blue to red. For text features, high TF-IDF values indicate low frequency of words in policy texts. Positive SHAP values indicate feature contribution towards advanced policy progression categories during training.}
\label{fig:TF-IDF_SHAP}
\end{figure}

Although BERT with Bayesian Ridge Regression performed best overall, text feature importance was not interpretable due to the sentence-level encoding of the policy texts. We therefore additionally investigated the explainability of the model using TF-IDF with CatBoost regression (RMSE = 0.18, R$^2$ = 0.2), which is compatible with SHAP analysis of text features and the results of which are presented in Fig.~\ref{fig:TF-IDF_SHAP}. The ``no party'' metadata feature directed policy prediction away from advanced progression stages, as indicated by the positive feature values associated with negative SHAP values. Similarly, high values for the ``MEPs'' metadata feature (share of seats held by political parties) directed policy prediction towards advanced progression stages. These results agree with the BERT feature permutation analysis, indicating that political party support is the strongest predictor of policy progression. 

In addition to metadata features, SHAP analysis enables investigation of the importance of text features. As seen in Fig.~\ref{fig:TF-IDF_SHAP}, the words ``environment'', ``europa'', and ``commission'' had higher importance when they occurred frequently across policies. This indicates that these words were useful in predicting policy advancement when they were prevalent across the dataset. Conversely, the words ``agreement'', ``climate'', and ``energy'' had positive SHAP values for high TF-IDF scores. This indicates that these words were useful when they were more prevalent only in specific policy categories of the texts dataset. Further analysis of the dataset found that, for example, ``climate'' frequently occurred in the ``Adopted/Completed'' policies, and ``energy'' in the ``Close to Adoption'' policies. 

\subsection{Implications and Broader Impact}
These findings demonstrate the potential of ML to support the analysis of climate policy texts and factors that contribute to policy progression. Increasing the transparency of policy progression can aid policy advocates, including NGOs and legislative rapporteurs, in understanding and developing legislative strategies.

Comparison of text representation methods demonstrated competitive performance of TF-IDF against more complex language models (Table \ref{tab:metadata_results}). This justifies the use of TF-IDF for text feature representation when computational power is a constraint. On text features alone, ClimateBERT demonstrated superior performance, reflecting its domain-specific pre-training and therefore its suitability in applications where metadata may be inaccessible or unavailable. 

Our results illustrate the trade-off between performance and explainability as more input features are included. While BERT methods performed best overall for the classification task, explainability was limited compared to TF-IDF methods due to the complexity of BERT text encoding. This is particularly relevant in the case of ClimateBERT, which has poor explainability if solely reliant on text features. Feature importance analysis illustrated the dominance of political party representation as a predictor for policy progression. This has significant implications, suggesting that the political context and support for a policy may be at least as important as the content of the policy itself, in order to progress policy through to adoption. However, consultation with policy experts is necessary in order to verify the interpretation of these results.

Although the dataset provides comprehensive coverage of policies available within the European Green Deal tracker up to January 2024, the overall sample size remains relatively small (165 policies). This constraint affects both the diversity of training examples and the statistical power of model evaluation. In particular, there is a pronounced class imbalance, with less than 5\% of policies labelled as ``Blocked'' or ``Withdrawn'' (Figure \ref{fig:ap_labels}). This class imbalance may affect advocacy use-cases, for example by providing false optimism. Therefore techniques such as undersampling or weighting should be considered alongside further application of this methodology to additional datasets to assess generalisability and robustness of the findings presented here.

Finally, while machine learning can enhance the transparency of policy processes, ethical considerations must remain central. Predictive models may inadvertently reflect systemic political biases (e.g. party representation). Such models should therefore be used as supporting tools, not decision-makers, and interpreted alongside expert judgement. Robustness across policy domains is also critical: applying these models to different legislative areas will help evaluate whether findings generalise or are specific to climate policy texts.

\section{Conclusion and Future Work}
This study highlights the value of ML in understanding climate policy progression within the European Green Deal. The integration of text and metadata features proved effective, and explainability techniques enhanced the interpretability of results: identifying metadata elements such as political party affiliation as critical predictors. Some of the identified features, such as party affiliation, are not modifiable with respect to the progression of specific policies. However, the predictive insights into institutional and political dynamics, which have been generated by this work, could help to inform advocacy strategisation and action towards shaping progression likelihood.

Future work will expand the dataset to incorporate policy design choices, such as instrument type, and explore temporal dynamics to improve model robustness. Additionally, ethical considerations such as data biases and transparency will remain central to developing actionable insights. This is particularly important as features like party affiliation may reflect systemic power imbalances that influence legislative outcomes. One possible extension of the presented work is to evaluate the predictive utility of the trained models on \emph{current} or ongoing policies whose final status is not yet known. This would assess whether the model can anticipate eventual progression categories. Such temporal evaluation is particularly relevant in the context of changing political landscapes following European elections, where rapporteur composition and party representation can shift significantly. While our current analysis focuses on retrospective prediction, applying these models prospectively would better reflect their potential for real-world policy monitoring and advocacy.

\begin{credits}
\subsubsection{\ackname} This study was supported by a Turing AI Fellowship grant (EP/V024817/1).

\subsubsection{\discintname}
The authors have no competing interests to declare that are
relevant to the content of this article.
\end{credits}

\bibliographystyle{splncs04}
\bibliography{iclr2025_conference}


\end{document}